\documentclass[10pt,twocolumn,letterpaper]{article}

\usepackage{cvpr}              % To produce the CAMERA-READY version
\usepackage{graphicx}
\usepackage{amsmath}
\usepackage{amssymb}
\usepackage{booktabs}
\usepackage[accsupp]{axessibility}
\graphicspath{ {./fig/} }

\usepackage[pagebackref,breaklinks,colorlinks]{hyperref}

\usepackage[capitalize]{cleveref}
\crefname{section}{Sec.}{Secs.}
\Crefname{section}{Section}{Sections}
\Crefname{table}{Table}{Tables}
\crefname{table}{Tab.}{Tabs.}

\def\cvprPaperID{1157} % *** Enter the CVPR Paper ID here
\def\confName{CVPR}
\def\confYear{2022}

\begin{document}

\title{Style-Based Global Appearance Flow for Virtual Try-On}

\author{Sen He, Yi-Zhe Song, Tao Xiang\\
Center for Vision, Speech and Signal Processing, University of Surrey\\ iFlyTek-Surrey Joint Research Centre on Artificial Intelligence\\
{\tt\small \{sen.he,y.song,t.xiang\}@surrey.ac.uk}
%\and
}

\twocolumn[{%
\renewcommand\twocolumn[1][]{#1}%
\maketitle
\begin{center}
\vspace{-7mm}
    \centering
    \includegraphics[width=1.0\textwidth]{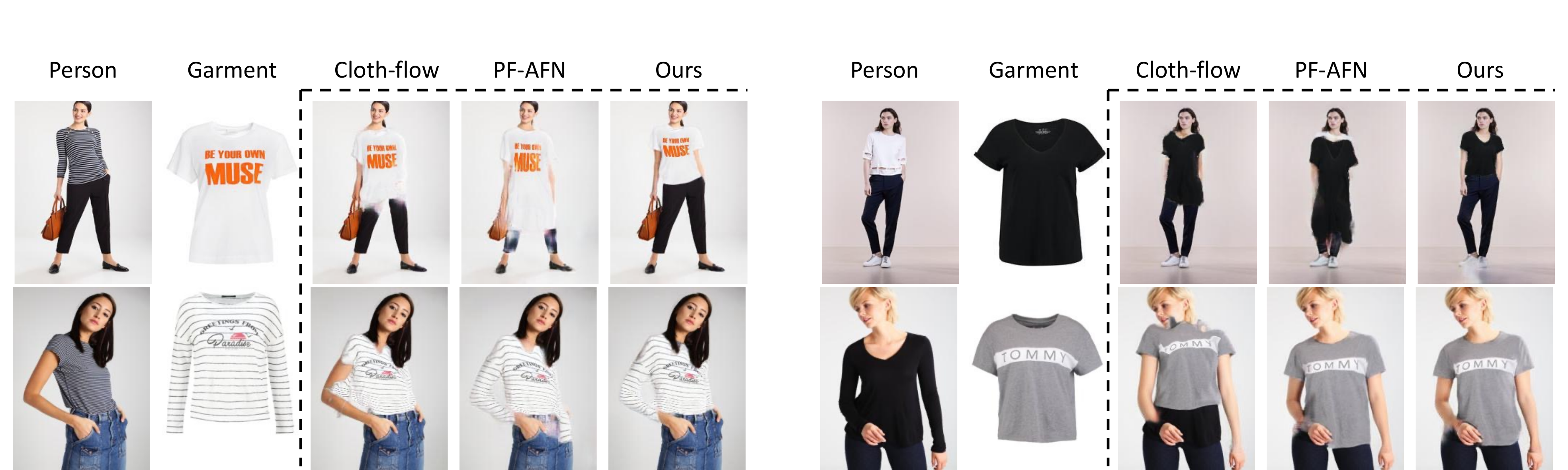}
    \captionof{figure}{Our global appearance flow based try-on model has a clear advantage over existing local flow based SOTA methods such as Cloth-flow \cite{han2019clothflow} and PF-AFN \cite{ge2021parser}, especially when there are large mis-alignment between reference and garment images (top row), and difficult poses/occlusions (bottom row).} %The middle row is a typical application scenario of virtual try-on where user wants to see the full body try-on effect. Existing local appearance flow estimation based methods, e.g., Clothe-flow \cite{han2019clothflow} and PF-AFN \cite{ge2021parser}, cannot handle these cases when there is a large misalignment between the person and the garment images. Our model, based on a novel global appearance flow method, is more robust in these cases.}
    \label{fig:fig1}
\end{center}%
}]

\maketitle

%%%%%%%%% ABSTRACT
\begin{abstract}
   Image-based virtual try-on aims to fit an in-shop garment into a clothed person image. To achieve this, a key step is garment warping which spatially aligns the target garment with the corresponding body parts in the person image.  Prior  methods typically adopt a local appearance flow estimation model. They are thus intrinsically susceptible to difficult body poses/occlusions and large mis-alignments between person and garment images (see Fig.~\ref{fig:fig1}). To overcome this limitation, a novel global appearance flow estimation model is proposed in this work. For the first time, a StyleGAN based architecture is adopted for appearance flow estimation. This enables us to take advantage of a global style vector to encode a whole-image context to cope with the aforementioned challenges. To guide the StyleGAN flow generator to pay more attention to local garment deformation, a flow refinement module is introduced to add local context. 
   %Our method is inspired from StyleGAN  for face manipulation where different style vectors can generate the same face at different viewpoints and different shapes. Note that both different viewpoints and shape deformations are explicit representations of the implicit appearance flows. We can therefore apply style modulation for appearance flow estimation. Different from previous methods, our method, via style modulation with a global style vector extracted from the global representations of the person and garment, first estimates a coarse appearance flow to warp the person image's feature and then refines the flow with local correspondence. 
   Experiment results on a popular virtual try-on benchmark show that our method achieves new state-of-the-art performance. It is particularly effective in a `in-the-wild' application scenario where the reference image is full-body resulting in a large mis-alignment with the garment image (Fig.~\ref{fig:fig1} Top). Code is available at: \url{https://github.com/SenHe/Flow-Style-VTON}.%Particularly, our method is more robust when the person is not well-positioned in the image, e.g., person's head is not aligned with the top of image.
   \vspace{-0.4cm}
\end{abstract}
%\vspace{-0.4cm}

%%%%%%%%% BODY TEXT
\section{Introduction}
The transition from offline in-shop retail to e-commerce has been accelerated by the recent pandemic caused lock downs.  In 2020, retail e-commerce sales worldwide amounted to 4.28 trillion US dollars and e-retail revenues are projected to grow to 5.4 trillion US dollars in 2022. However, when it comes to fashion, one of key offline experiences missed by the on-line shoppers is the changing room where a garment item can be tried-on. To reduce the return cost for the online retailers and give shoppers the same offline experience online, image-based virtual try-on (VTON) has been studied intensively recently  \cite{han2018viton,wang2018toward,yu2019vtnfp,yang2020towards,issenhuth2020not,ge2021disentangled, wang2020down, han2019clothflow, ge2021parser, lewis2021tryongan}.

A VTON model aims to  fit an in-shop garment into a person image. 
A key objective of a VTON model is to align the in-shop garment with the corresponding body parts in the person image. This is due to the fact that the in-shop garment is usually not spatially aligned with the person image (see Fig.~\ref{fig:fig1}). Without the  spatial alignment, directly applying advanced detail-preserving image to image translation models \cite{ronneberger2015u, isola2017image} to fuse the texture in person image and garment image will result in unrealistic effect in the generated try-on image, especially in the occluded and misaligned regions.

Previous methods address this alignment problem through garment warping, i.e., they first warp the in-shop garment, which is then concatenated with the person image and fed into an image to image translation model for the final try-on image generation.  Many of them \cite{han2018viton,wang2018toward,yu2019vtnfp,yang2020towards,issenhuth2020not,ge2021disentangled} adopt a Thin Plate Spline (TPS) \cite{duchon1977splines} based on the warping method,  exploiting the correlation between features extracted from the person and garment images. However, as analyzed in previous works \cite{yang2020towards, han2019clothflow,chopra2021zflow}, TPS has limitations in handling complex warping, e.g., when different regions in the garment require different deformations. As a result, recent SOTA methods \cite{han2019clothflow, ge2021parser} estimate dense appearance flow \cite{zhou2016view} to warp the garment. This involves training a network to predict the dense appearance flow field representing the deformation required to align the garment with the corresponding body parts. %Compared to TPS based warping, appearance flow based warping has a larger degree of freedom and is more superior in modeling complex deformation. 

However, existing appearance flow estimation methods are limited in accurate garment warping due to the lack of global context. More specifically, all existing methods are based on local feature's correspondence, e.g., local feature concatenation or correlation\footnote{It is worth noting that the tensor correlation methods \cite{ge2021parser,dosovitskiy2015flownet, ilg2017flownet} have the potential to reach global receptive field. However, its computation grows quadratically with respect to the input size. To make it tractable, its actual implementation is still based on limited local neighborhoods.}, developed for optical flow estimation \cite{dosovitskiy2015flownet, ilg2017flownet}. To estimate the appearance flow, they make the unrealistic assumption that the corresponding regions from the person image and the in-shop garment are located in the same local receptive filed of the feature extractor. When there is a large mis-alignment between the garment and corresponding body parts (Fig.~\ref{fig:fig1} Top), current appearance flow based methods will deteriorate drastically and generate unsatisfactory results. Lacking a global context also make existing  flow-based VTON methods vulnerable to difficult poses/occlusions (Fig.~\ref{fig:fig1} Bottom) when correspondences have to be searched beyond a local neighborhood. This severely limits the use of these methods `in-the-wild', whereby a user may have a full-body picture of herself/himself as the person image to try-on multiple garment items (e.g., top, bottom, and shoes).  

%the person in the image is not well positioned and the corresponding regions, e.g., neck in the person image and collar in the in-shop garment, are not in the same receptive filed at all, current appearance flow based methods will deteriorate and generate unsatisfactory results (see examples in Fig.~\ref{fig:fig1},\ref{fig:fig5}).

 To overcome this limitation, a novel global appearance flow estimation model is proposed in this work. Specifically, for the first time,  a StyleGAN \cite{karras2019style,karras2020analyzing} architecture for dense appearance flow estimation. This differs fundamentally from existing methods \cite{ge2021parser,han2019clothflow, dosovitskiy2015flownet, ilg2017flownet} which employ a U-Net \cite{ronneberger2015u} architecture to preserve local spatial context. Using a global style vector extracted from the whole reference and garment images makes  it easy for our model to capture global context. However, it also raises an important question: can it capture local spatial context crucial for local alignments? After all, a single style vector seemingly has lost local spatial context. To answer this question, we first  note that StyleGAN has been successfully applied to local face image manipulation tasks, where different style vectors can generate the same face at different viewpoints \cite{shen2021closed} and different shapes \cite{or2020lifespan,he2021disentangled}. This suggests that a global style vector does have local spatial context encoded. However, we also note that the vanilla StyleGAN architecture \cite{karras2019style, karras2020analyzing}, though much more robust against large mis-alignment and difficult poses/occlusions compared to U-Net, is weaker when it comes to local deformation modeling. We therefore introduce a local flow refinement module in the existing StyleGAN generator to have the better of both worlds.  %   Note that different viewpoints and different shapes are explicit representation of the implicit appearance flow. If the style vector can manipulate the explicit representation, it should be able to predict the implicit appearance flow. Therefore, we apply style modulation into VTON for in-shop garment warping. Importantly, the style vector used to predict the appearance flow is a global representation of the person and garment, it is therefore more robust with respect to various person images and thus more practical in real-world scenarios of VTON.

Concretely, our StyleGAN-based warping module ($\mathcal{W}$ in Fig.~\ref{fig:fig_archi}) consists of stacked warping blocks that takes as inputs  a global style vector, garment features and person features. The global style vector is  computed from  the lowest resolution feature maps of the person image and the in-shop garment for global context modeling. In each warping block in the generator, the global style vector is used to modulate the feature channels which takes in the corresponding garment feature map to estimate the appearance flow. To enable our flow-estimator to model the fine-grained local appearance flow, e.g., the arm and hand regions in Fig.~\ref{fig:fig4}, in each warping block on top of the style based appearance flow estimation part, we introduce a refinement layer. This refinement layer first warps the garment feature map, which is subsequently  concatenated with the person feature map at the same resolution and then used to predict the local detailed appearance flow. %Note that after the warping by the flow predicted by the global style vector, the warped garment feature map is already coarsely aligned with the corresponding part in person image feature map. The local refinement layer therefore won't suffer from the poorly positioned person image as in previous local correspondence based appearance flow based methods \cite{han2019clothflow, ge2021parser}. 

\textbf{The contributions} of this work are as follow: (1) We propose a novel style-based appearance flow method to warp the garment in virtual try-on. This global flow estimation approach makes our VTON model much robust against large mis-alignments between person and garment images. This makes our method more applicable to `in-the-wild' application where a full-body person image with natural poses is used (see in Fig.~\ref{fig:fig1}). (2) We conduct extensive experiments to validate our method, demonstrating clearly that it is superior to existing state-of-the-art alternatives.

\section{Related Work}
\paragraph{Image based virtual try-on} Image based (2D) VTON can be categorized into parser-based methods and parser-free methods. Their main difference is whether an off-the-shelf human parser\footnote{Sometimes, pre-trained pose \cite{cao2017realtime} and densePose \cite{guler2018densepose} detection models are also used in a parser based model.} is required in the inference stage.

Parser-based methods apply a human segmentation map to mask the garment region in the input person image for warping parameter estimation. The masked person image is concatenated with the warped garment and then fed into a generator for target try-on image generation. Most methods \cite{han2018viton,wang2018toward,yu2019vtnfp,han2019clothflow,yang2020towards,ge2021disentangled} apply a pre-trained human parser \cite{gong2017look} to parse the person image into several pre-defined semantic regions, e.g., head, top, and pants. For better try-on image generation, \cite{yang2020towards} also transforms the segmentation map to  match the target garment. The transformed parsing result, together with the warped garment and the masked person image are used for final try-on image generation. The reliance on a parser make these methods sensitive to bad human parsing results \cite{issenhuth2020not, ge2021parser} which inevitably  lead to inaccurate  warping and try-on results.

In contrast, parser-free methods \cite{issenhuth2020not, ge2021parser}, in the inference stage, only takes as inputs  the person image the garment image. They   are designed specifically to eliminate the negative effects induced by the bad parsing results. Those methods usually first train a parser-based teacher model and then distill a parser-free student model. \cite{issenhuth2020not} proposed a pipeline which distills the garment warping module and try-on generation network using paired triplets. \cite{ge2021parser} further improved \cite{issenhuth2020not} by introducing cycle-consistency for better distillation.

Our method is also a parser free method. However, our method focuses on the design of the garment warping part, where we propose a novel global appearance flow based garment warping module.
\vspace{-0.4cm}

\paragraph{3D virtual try-on} Compared to image based VTON, 3D VTON provides better try-on experience (e.g., allowing being viewed with arbitrary views and poses),  yet is also more challenging. Most 3D VTON works \cite{bhatnagar2019multi, mir2020learning} rely on 3D parametric human body models \cite{loper2015smpl} and need scanned 3D datasets for training.   Collecting large scale 3D datasets is expensive and laborious, thus posing a constraint on the scalability of a 3D VTON model. To overcome this problem, recently \cite{zhao2021m3d} applied non-parametric dual human depth model \cite{gabeur2019moulding} for monocular to 3D VTON. However, existing 3D VTON still generate inferior texture details compared to the 2D methods.  

\vspace{-0.4cm}
\paragraph{StyleGAN for image manipulation}   StyleGAN \cite{karras2019style, karras2020analyzing} has revolutionized the research on image manipulation \cite{shen2020interfacegan, yang2021semantic, or2020lifespan} lately.  Its successful application on the image manipulation tasks often thanks to its suitability in learning a highly disentangled latent space. Recent efforts have been  focused on  unsupervised latent semantics discovery \cite{shen2021closed,cherepkov2021navigating, tzelepis2021warpedganspace}. \cite{lewis2021tryongan} applied pose conditioned StyleGAN for virtual try-on. However, their model cannot preserve garment details and is slow during inference.

The design of our garment warping network is inspired from StyleGAN in image manipulation, especially its super performance in shape deformation \cite{shen2021closed, or2020lifespan}. Instead of using style modulation to generate the warped garment, we use style modulation to predict the implicit appearance flow which is then used to warp the garment via sampling. This design is much more suited to garment detail-preserving compared to \cite{lewis2021tryongan}.
\vspace{-0.4cm}
\paragraph{Appearance flow} In the context of VTON, appearance flow was first introduced by \cite{han2019clothflow}. Since then, it has gained more attention and adopted by recent state-of-the-art VTON models \cite{ge2021parser, chopra2021zflow}. Fundamentally, appearance flow is used as a sampling grid for garment warping, it is thus information lossless and superior in detail preserving. 
Beyond VTON, appearance flow is also popular in other tasks. \cite{zhou2016view} applied it for novel view synthesis. \cite{ren2020deep, albahar2021pose} also applied the idea of appearance flow to warp the feature map for person pose transfer.
 Different from all these existing appearance flow estimation methods, our method, via style modulation, applies a global style vector to estimate the appearance flow. Our method is thus intrinsically superior in its ability to coping with large mis-alignments.

\section{Methodology}
\begin{figure*}[t!]
    \centering
    \includegraphics[width=1.0\textwidth]{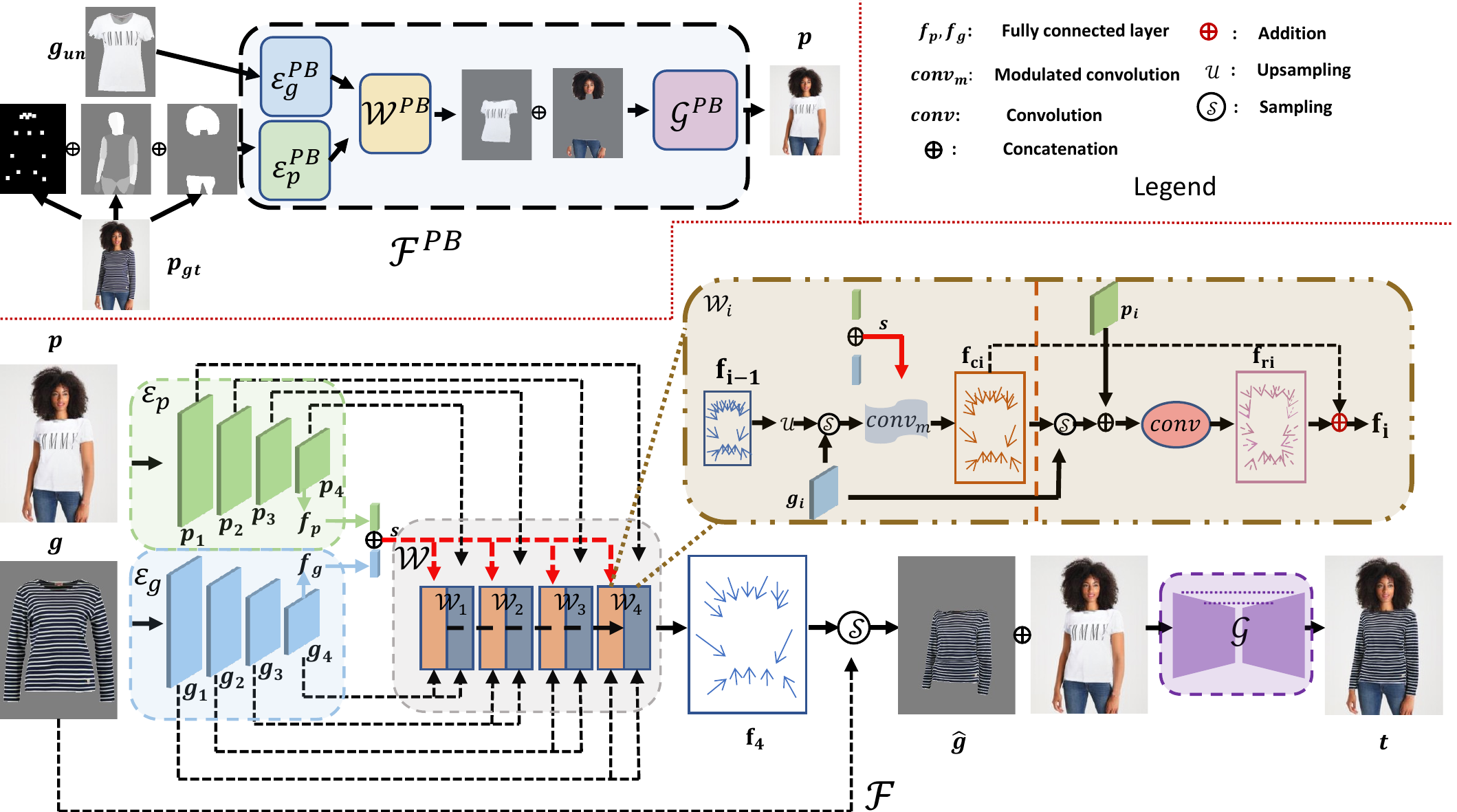}
    \caption{A schematic of our framework. The pre-trained parser based model $\mathcal{F}^{PB}$ generates an output image as the input of parser free model $\mathcal{F}$. The two feature extractors in $\mathcal{F}$ extract the feature of person image and garment image, respectively. A style vector is extracted from the lowest resolution feature maps from person image and the garment image. The warping module takes in the style vector and feature maps from the person image and garment image, and output an appearance flow map. The appearance flow is then used to warp the garment. Finally, the warped garment is concatenated with person image and fed into the generator to generate the target try-on image. Note that $\mathcal{F}_{PB}$ is only used during training.}
    \vspace{-0.4cm}
    \label{fig:fig_archi}
\end{figure*}

\subsection{Problem definition}

Given a person image ($p \in \mathbb{R} ^ {3\times H \times W}$) and an in-shop garment image ($g \in \mathbb{R} ^ {3 \times H \times W}$), the goal of virtual try-on is to generate a try-on image ($t \in \mathbb{R} ^ {3 \times H \times W}$) where the garment in $g$ fits to the corresponding parts in $p$. In addition, in the generated $t$, both details from $g$ and non-garment regions in $p$ should be preserved. In other words, the same person in $p$ should appear unchanged in $t$ except now wearing $g$.

To eliminate the negative effect of inaccurate human parsing, our proposed model ($\mathcal{F}$ in Fig.~\ref{fig:fig_archi}) is designed to be a parser-free model. Following the strategy adopted by existing parser-free models \cite{issenhuth2020not, ge2021parser}, we first pre-train a parser-based model ($\mathcal{F}^{PB}$). It is then used as a teacher for knowledge distillation to help train the final parser-free model $\mathcal{F}$. Both $\mathcal{F}$ and $\mathcal{F}^{PB}$ consist of three parts, i.e., two feature extractors ($\mathcal{E}_{p}^{PB}$, $\mathcal{E}_{g}^{PB}$ in $\mathcal{F}^{PB}$ and $\mathcal{E}_{p}$, $\mathcal{E}_{g}$ in $\mathcal{F}$), warping module ($\mathcal{W}^{PB}$ in $\mathcal{F}^{PB}$ and $\mathcal{W}$ in $\mathcal{F}$), and a generator ($\mathcal{G}^{PB}$ in $\mathcal{F}^{PB}$ and $\mathcal{G}$ in $\mathcal{F}$). Each of them will be detailed in the following sections.

\subsection{Pre-training a parser-based model}

As per standard in existing parser-free models \cite{issenhuth2020not, ge2021parser}, a parser-based model $\mathcal{F}^{PB}$ is first trained. It is used in two ways in the subsequent training of the proposed parser-free model $\mathcal{F}$: (a) to generate  person image ($p$) to be used by $\mathcal{F}$ as input and (b) to supervise the training of $\mathcal{F}$ via knowledge distillation.

Concretely, $\mathcal{F}_{PB}$ takes as inputs the semantic representation (segmentation map\footnote{The garment region in the segmentation map is flipped as background region}, keypoint pose and dense pose) of a real person image ($p_{gt} \in \mathbb{R}^{3 \times H \times W}$) in the training set and an unpaired garment ($g_{un} \in \mathbb{R}^{3 \times H \times W}$). The output of $\mathcal{F}_{PB}$ is the image $p$ where the original person is wearing $g_{un}$. $p$ will serve as the input for $\mathcal{F}$ during training.  This design, according to \cite{ge2021parser}, benefits from the fact that we now have  paired person image $p_{gt}$ and garment image $g$ in $p_{gt}$ to train the parser-free model $\mathcal{F}$, that is:
\begin{equation}
    \mathcal{F}^{*} = \underset{\mathcal{F}}{\text{arg min}} \lVert t - p_{gt}\rVert,
\end{equation}
where $t = \mathcal{F}(p,g)$ is the generated try-on image from $\mathcal{F}$. Note that $\mathcal{F}^{PB}$ is only used during the training of $\mathcal{F}$.

\subsection{Feature extraction}

We apply two convolutional encoders ($\mathcal{E}_{p}$ and $\mathcal{E}_{g}$) to extract the features of $p$ and $g$. Both $\mathcal{E}_{p}$ and $\mathcal{E}_{g}$ share the same architecture, composed of stacked residual blocks. The extracted features from $\mathcal{E}_{p}$ and $\mathcal{E}_{g}$ can be represented as $\{p_{i}\}_{1}^{N}$ and $\{g_{i}\}_{1}^{N}$ ($N=4$ in Fig.~\ref{fig:fig_archi} for simplicity), where $p_{i} \in \mathbb{R}^{c_{i} \times h_{i} \times w_{i}}$ and $g_{i} \in \mathbb{R}^{c_{i} \times h_{i} \times w_{i}}$ are the feature maps extracted from the corresponding residual block in $\mathcal{E}_{p}$ and $\mathcal{E}_{g}$, respectively. The extracted feature maps will be used in $\mathcal{W}$ to predict the appearance flow.

\subsection{Style based appearance flow estimation}

The main novel component of the proposed model is a  style-based global appearance flow estimation module. Different from previous methods that estimate appearance flow based on local feature correspondence \cite{han2019clothflow, ge2021parser}, originally proposed in optical flow estimation \cite{dosovitskiy2015flownet, ilg2017flownet}, our method, based on a global style vector, first estimates a coarse appearance flow via style modulation and then refine the predicted coarse appearance flow based on local feature correspondence. 

As illustrated in Fig.~\ref{fig:fig_archi}, our warping module ($\mathcal{W}$) consists of N stacked warping blocks ($\{\mathcal{W}_{i}\}_{1}^{N}$), each block is composed of a style-based appearance flow prediction layer (orange rectangle) and a local correspondence based appearance flow refinement layer (blue rectangle). Concretely, we first extract a global style vector ($s \in \mathbb{R}^{c}$) using the features output from the  $N^{th}$ (final) blocks of $\mathcal{E}_{p}$ and $\mathcal{E}_{g}$, denoted as $p_{N}$ and $g_{N}$, as:
\begin{equation}
    s = [f_{p}(p_{N}), f_g(g_{N})],
\end{equation}
where $f_{p}$ and $f_{g}$ are  fully connected layers, and $[\cdot,\cdot]$ denotes concatenation. Intrinsically, the extracted global style vector $s$\footnote{Intuitively, $s = f_{p}(p_{N})$ is enough to generate the appearance flow. But we empirically found that $s = [f_{p}(p_{N}), f_g(g_{N})]$ yields better results.} contains the global information of the person and garment, e.g., position, structure, etc. Similar to style based image manipulation \cite{shen2020interfacegan,shen2021closed, or2020lifespan,he2021disentangled}, we expect the global style vector $s$ capture the required deformation for warping $g$ into $p$. It is thus used for  style modulation in a StyleGAN style generator for estimating a appearance flow field.  

More specifically, in the style-based appearance flow prediction layer of each block $\mathcal{W}_{i}$, we apply style modulation to predict a coarse flow:
\begin{equation}\label{eq:2}
    \mathbf{f_{ci}} = conv_m(\mathcal{S}(g_{N+1-i}, \mathcal{U}(\mathbf{f_{i-1}})),s),
\end{equation}
where $conv_{m}$ denotes modulated convolution \cite{karras2019style}, $\mathcal{S}(\cdot,\cdot)$ is the sampling operator, $\mathcal{U}$ is the upsampling operator, and $\mathbf{f_{i-1}} \in \mathbb{R}^{2 \times h_{i-1} \times w_{i-1}}$ is the predicted flow from last warping block. Note that the first block $\mathcal{W}_{1}$ in $\mathcal{W}$ only takes in the lowest resolution garment feature map and the style vector, i.e., $\mathbf{f_{c1}} = conv_{m}(g_{N}, s)$. As can be seen from Equation~\ref{eq:2}, the predicted $\mathbf{f_{ci}}$ depends on the garment feature map and the global style vector. It thus has a global receptive field and is capable to cope with large mis-alignments between the garment and  person images. However, as the style vector $s$ is a global representation, as a trade-off, it has a limited ability to accurately estimate the local fine-grained appearance flow (as shown in Fig.~\ref{fig:fig4}). The coarse flow is thus in need of a local refinement.

To refine $\mathbf{f_{ci}}$, we introduce a local correspondence based appearance flow refinement layer in  each block $\mathcal{W}_{i}$. It aims to estimate a local fine-grained appearance flow: 
\begin{equation}\label{eq:3}
    \mathbf{f_{ri}} = conv([\mathcal{S}(g_{N+1-i}, \mathbf{f_{ci}}),p_{N+1-i}]),
\end{equation}
where $\mathbf{f_{ri}}$ is the predicted refinement flow, and $conv$ denotes convolution. Fundamentally, the refinement layer estimates the refinement flow through the local correspondence, i.e., the correspondence between warped person features and garment feature in the same receptive field. Note that after the warping by $\mathbf{f_{ci}}$, we can assume that the corresponding regions/features in $g_{N+1-i}$ and $p_{N+1-i}$ are now located in the same receptive field. Therefore, we can apply the local correspondence used in previous works \cite{han2019clothflow,ge2021parser} to predict the local fine-grained appearance flow.

Finally, we add the coarse flow and the local fine-grained appearance flow together as the output of each warping block:

\begin{equation}\label{eq:4}
    \mathbf{f_{i}} = \mathbf{f_{ci}} + \mathbf{f_{ri}}.
\end{equation}

The predicted appearance flow $\mathbf{f_{N}}$ from the last block in $\mathcal{W}$ is used to warp the garment:

\begin{equation}
    \hat{g} = \mathcal{S}(g, \mathbf{f_{N}}).
\end{equation}
And the warped garment $\hat{g}$ is then concatenated with the person image and fed into a generator for target try-on image generation:
\begin{equation}
    t = \mathcal{G}([\hat{g}, p]).
\end{equation}
The generator $\mathcal{G}$ has an encoder-decoder architecture with skip connections in between. We follow the designs in \cite{isola2017image,zhu2017unpaired} that have been proven to be effective in texture detail preservation.

\subsection{Learning objectives}

To train our model, we first apply a perceptual loss \cite{johnson2016perceptual} between the output of $\mathcal{F}$ and the ground truth person image $p_{gt}$:

\begin{equation}
    L_{p} = \sum_{i} \lVert \phi_{i}(t) - \phi_{i}(p_{gt}) \rVert,
\end{equation}
where $\phi_{i}$ is the $i^{th}$ block of the pre-trained VGG network \cite{simonyan2014very}.

To supervise the training of the warping model $\mathcal{W}$, we apply a loss on the warped garment:
\begin{equation}
    L_{g} = \lVert \hat{g} - m_{g} \cdot p_{gt}\rVert,
\end{equation}
where $m_{g}$ is the garment mask of $p_{gt}$ predicted by an off-the-shelf human parsing model.

As per standard in previous appearance flow methods \cite{ge2021parser,han2019clothflow}, we also apply a smoothness regularization on the predicted flow from each block in $\mathcal{W}$:
\begin{equation}
    L_{R} = \sum_{i} \lVert \nabla \mathbf{f_{i}} \rVert,
\end{equation}
where $\lVert \nabla \mathbf{f_{i}} \rVert$ is the generalized charbonnier loss function \cite{sun2014quantitative}.

As the inputs (segmentation map, keypoint pose and dense pose) to the parser-based person encoder ($\mathcal{E}^{PB}$) contain more semantic information than those of the parser-free model $\mathcal{F}$ (person image), we apply a distillation loss to guide the learning of person encoder $\mathcal{E}_{p}$ in $\mathcal{F}$:
\begin{equation}
    L_{D} = \sum_{i} \lVert p_{i}^{PB} - p_{i}\rVert,
\end{equation}
where $p_{i}^{PB}$ is the output feature map from $i^{th}$ block in the person encoder $\mathcal{E}_{p}^{PB}$ in the pre-trained parser based model $\mathcal{F}_{PB}$.

The overall learning objective is:

\begin{equation}
    L = \lambda_{p}L_{p} + \lambda_{g}L_{g} + \lambda_{R}L_{R} + \lambda_{D}L_{D},
\end{equation}
where $\lambda_{p}$, $\lambda_{g}$ ,$\lambda_{R}$ and $\lambda_{D}$ denote the hyperparameters for balancing the four objectives.

\section{Experiments}

\paragraph{Datasets} We experiment our model on the VITON dataset\footnote{The usage of the dataset has been permitted by the author in \cite{han2018viton}.} \cite{han2018viton}. It is the most popular dataset used in previous VTON works. VITON contains a training set containing $14,221$ image pairs\footnote{Each pair means a person image and the image of garment on the person.} and a testing dataset of $2,032$ pairs. Both person and garment images are of the resolution $256 \times 192$. 

We also create a testing dataset, denoted by augmented VITON, to evaluate model's robustness to the random positioned person image (see example in Fig.~\ref{fig:fig3_aug}) with larger mis-alignments with the garment images in the original dataset. As most testing person images in VITON are well positioned such that the person image and the garment are well pre-aligned (e.g., most corresponding regions in the person image and garment image are roughly located in the same receptive field), it is not suited for this evaluation. Concretely, the augmented VITON dataset is created by randomly augmenting the testing person image in VITON via shifting and zooming in/out. In particular, we randomly augment 1/3 testing person images in VITON by shifting the person's position in the image and randomly augment another 1/3 test images in VITON by zooming in/out the person in the image and keep another 1/3 testing images unchanged. When evaluated on this dataset,  all compared models are trained with person image augmentation.
%\vspace{-0.4cm}

\paragraph{Implementation details} Our model is implemented in PyTorch. We train our model with a single Nvidia RTX 2080-Ti GPU. We set the batch size as 4 and train the model with 100 epochs. We train the model with Adam optimizer \cite{kingma2014adam}. The initial learning rate is set to $5e-4$ which is linearly decayed after 50 epochs. Each residual block in $\mathcal{E}_{p}$ and $\mathcal{E}_{g}$ is followed by a pooling layer to reduce the spatial dimension. We set $N=5$ and $c=256$ in the implementation. We will release the code upon the acceptance of this work.
\vspace{-0.4cm}
\paragraph{Evaluation metrics and baselines} We evaluate our model both automatically and manually. In the automatic evaluation, as per standard in VTON, we evaluate model performance using structure similarity (SSIM) \cite{wang2004image} and Fr\'{e}chet Inception Distance (FID)\cite{heusel2017gans}. According to \cite{rosca2017variational,ge2021parser}, inception score (IS) \cite{salimans2016improved} is not suitable to evaluate VTON images, we thus do not adopt it in the evaluation. In the manual (subjective) evaluation, we run perceptual study on
Amazon Mechanical Turk (AMT) to compare the quality
of the generated try-on images from different models. Given an  input person image, a garment image and the generated try-on image from two models, the AMT workers were asked to vote which generated try-on image is better. Each AMT worker was randomly allocated 100 images to compare two models. 15 AMT workers participated in the evaluation for all models comparison.

\begin{figure*}[t!]
    \centering
    \includegraphics[width=0.9\textwidth]{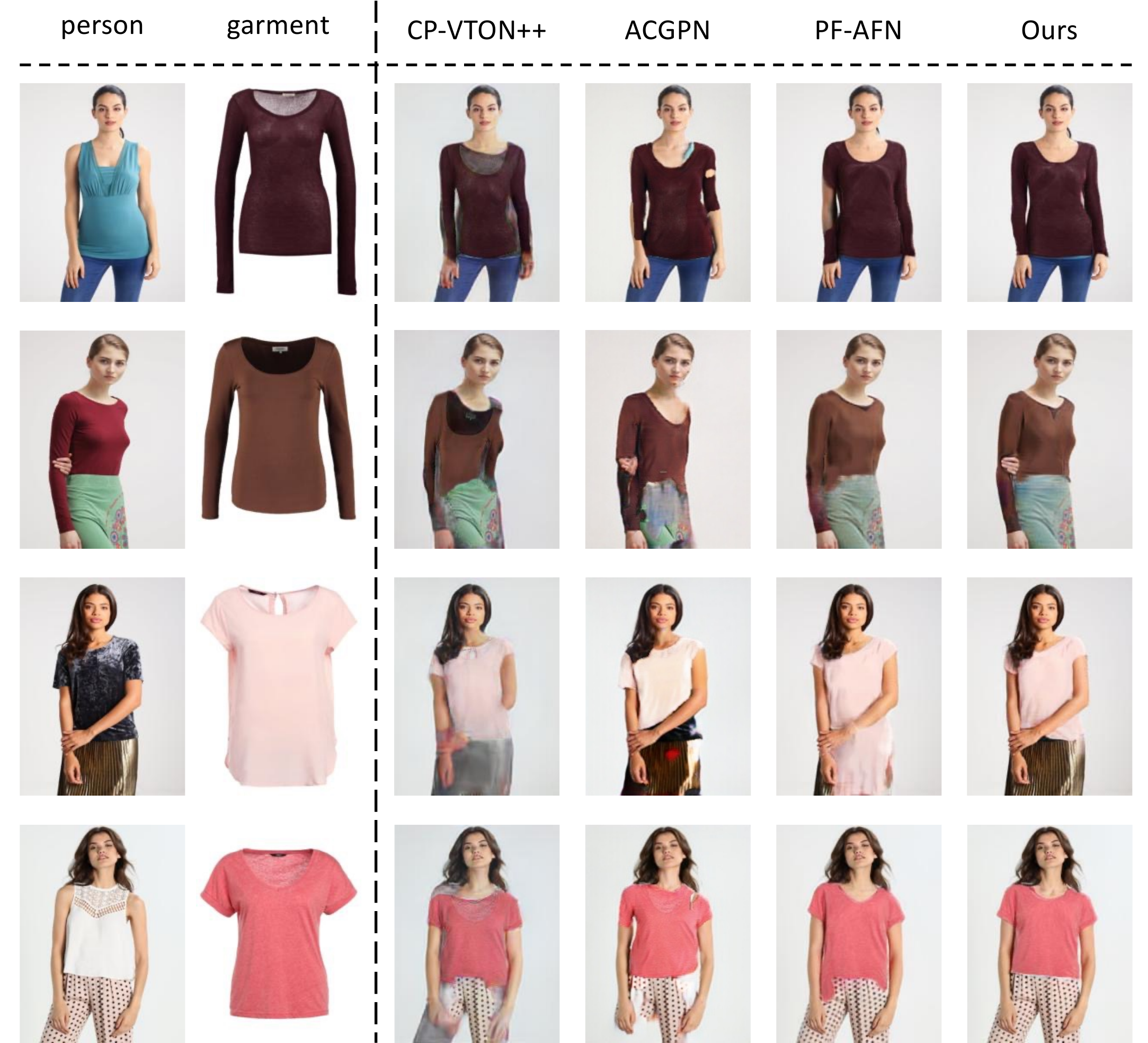}
    \caption{Qualitative results from different models (CP-VTON++ \cite{minar2020cp}, ACGPN \cite{yang2020towards}, PF-AFN \cite{ge2021disentangled} and ours) on VITON testing dataset.}
    \vspace{-0.0cm}
    \label{fig:fig3}
\end{figure*}

\begin{figure*}[t!]
    \centering
    \includegraphics[width=0.9\textwidth]{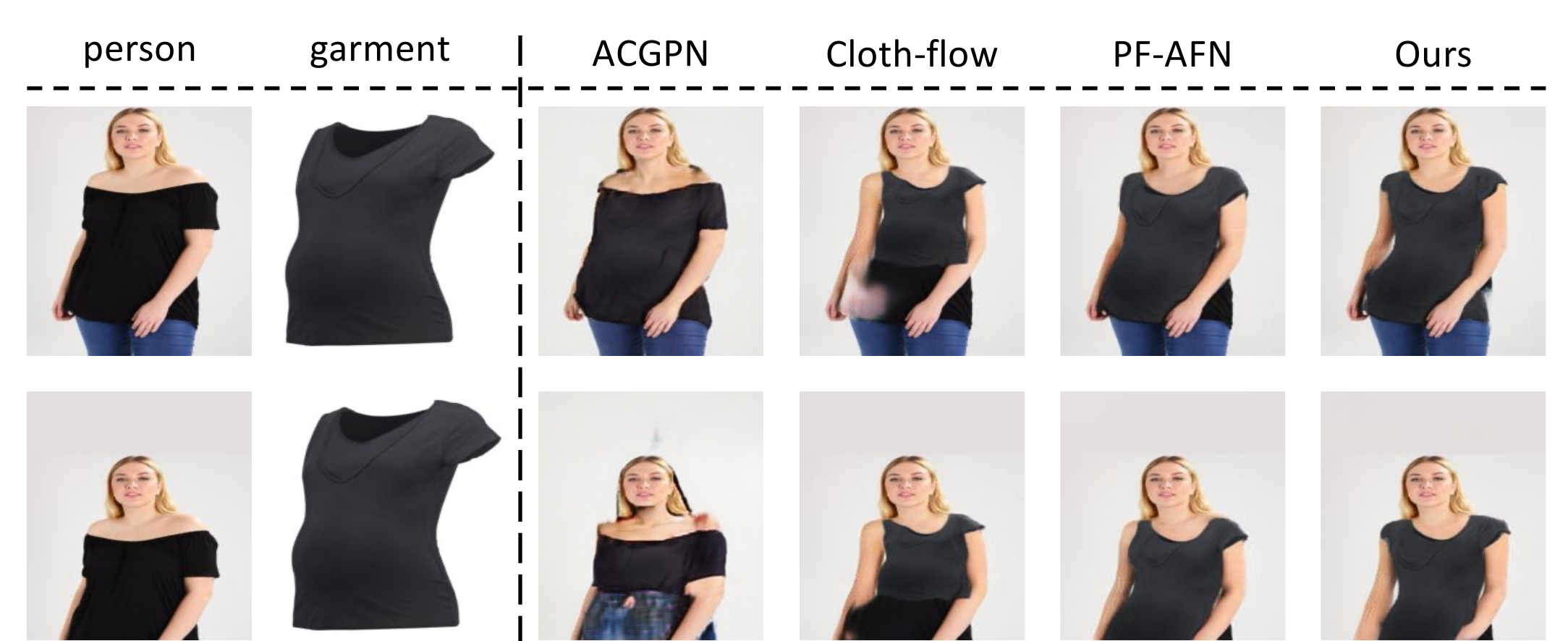}
    \caption{Illustrating different VTON models' robustness to the randomly positioned person image. First row uses original person image as input. And second row uses vertically shifted person image as input. ACGPN \cite{yang2020towards}, Cloth-flow \cite{han2019clothflow}, PF-AFN \cite{ge2021parser}.}
    \vspace{-0.4cm}
    \label{fig:fig3_aug}
\end{figure*}

We compare our methods with other parser-based methods VTON \cite{han2018viton}, CP-VTON \cite{wang2018toward}, Cloth-flow \cite{han2019clothflow}, CP-VTON++ \cite{minar2020cp}, ACGPN \cite{yang2020towards}, DCTON \cite{ge2021disentangled} and ZFlow \cite{chopra2021zflow}. We also compare with the SOTA parser-free method PF-AFN \cite{ge2021parser}. 
\vspace{-0.4cm}
\paragraph{Main results} The quantitative results on VITON testing dataset are  shown in Table~\ref{tab:tab1}. It can be seen that our model achieves new state-of-the-art performance. Importantly, given the already low FID score (10.09) achieved by prior SOTA method PF-AFN,  our method can further decrease it by $11.9\%$. In the meanwhile, the following observations can be made from  Table~\ref{tab:tab1}. (1) Appearance flow based warping methods generally perform better than TPS based warping methods. (2) Although it takes more training time, parser-free methods are much better than parser-based methods. Our model, benefiting from the proposed novel global appearance flow estimation method, outperforms the previous SOTA parser-free methods (PF-AFN \cite{ge2021parser} and Cloth-flow \cite{han2019clothflow}) on all evaluation metrics.  
The human evaluation results are shown in Table~\ref{tab:tab2}. The result is consistent with that in Table~\ref{tab:tab1}. Our model outperforms all compared models with more than $10\%$ preference rate. The qualitative results from different models are illustrated in Fig.~\ref{fig:fig3}. Overall, our method generates better try-on images. For example, the hard pose and occlusion in second and third rows. 

The quantitative results on augmented testing dataset are shown in Table~\ref{tab:tab3}. As can be seen that our model again performs best on the augmented VITON testing dataset. Importantly, all other models' performance drops dramatically. And our model can still maintain the performance (SSIM score) compared to that on the original VITON testing dataset. The qualitative examples are illustrated in Fig.~\ref{fig:fig3_aug}. Only our model can generate consistent (e.g., the garment's left sleeve) and high quality try-on images given the large mis-alignments.
\vspace{-0.4cm}

\begin{table}[t]
    \centering
    \small\addtolength{\tabcolsep}{-1pt}
    \begin{tabular}{c|c|c|c|c}
    \toprule
         Methods & Warping& Parser& SSIM $\uparrow$& FID$\downarrow$\\
         \midrule
         VTON \cite{han2018viton}& TPS& Y &0.74& 55.71\\
         CP-VTON \cite{wang2018toward}&TPS&Y &0.72 &24.45\\
         CP-VTON++ \cite{minar2020cp}& TPS&Y&0.75& 21.04\\
         Cloth-flow\cite{han2019clothflow}&AF&Y&0.84& 14.43\\ 
         ACGPN\cite{yang2020towards}& TPS&Y&0.84& 16.64\\
         DCTON\cite{ge2021disentangled}&TPS&Y& 0.83& 14.82\\
         PF-AFN\cite{ge2021parser} &AF&N&0.89 &10.09\\
         Zflow \cite{chopra2021zflow}&AF&Y&0.88&15.17\\ 
         $\text{Cloth-flow}^{\star}$\cite{han2019clothflow}&AF&N&0.89& 10.73\\
         \midrule
         Ours&AF&N&\textbf{0.91}&\textbf{8.89}\\
         \bottomrule
    \end{tabular}
    \caption{Quantitative results of different models on VITON. Warping represents the warping methods used in different models. Parser indicates whether human parser is used in the model during inference. TPS: Thin Plate Spline. AF: Appearance Flow. $\star$: re-trained with parser free training paradigm.}
    \label{tab:tab1}
\end{table}

\begin{table}[t]
    \centering
    \begin{tabular}{c|c}
    \toprule
         Compared methods & preference rate\\
         \midrule
         
         CP-VTON++ \cite{minar2020cp}& 12.7\% / 87.3\%\\
         
         ACGPN\cite{yang2020towards}& 20.2\% / 79.8\%\\
         
         $\text{Cloth-flow}^{\star}$\cite{han2019clothflow}& 38.5\% / 61.5\%\\
         
         AF-PFN\cite{ge2021parser}&43.2\% / 56.8\%\\
         
         \bottomrule
    \end{tabular}
    \vspace{-0.2cm}
    \caption{The preference rate comparing other models against our model (other models/our model) in human evaluation.}
    \label{tab:tab2}
    \vspace{-0.4cm}
\end{table}

\begin{table}[t]
    \centering
    \begin{tabular}{c|c|c|c}
    \toprule
         Methods & SSIM $\uparrow$& FID$\downarrow$&$\triangledown_{\text{SSIM}}$/$\triangledown_{\text{FID}}$\\
         \midrule
         
         ACGPN &0.81&20.75&0.003/4.11\\
         $\text{Cloth-flow}^{\star}$\cite{han2019clothflow}&0.86& 13.05&0.003/2.96\\ 
         AF-PFN\cite{ge2021parser} &0.87 &12.19&0.002/2.10\\
         \midrule
         Ours&\textbf{0.91}& \textbf{9.91}&\textbf{0/1.02}\\
         \bottomrule
    \end{tabular}
    \caption{Quantitative results of different models on augmented VITON and their relative performance drop ($\triangledown_{\text{SSIM}}$/$\triangledown_{\text{FID}}$) compared to the standard VITON testing dataset.}
    \vspace{-0.4cm}
    \label{tab:tab3}
\end{table}

\begin{figure}[h!]
    \centering
    \includegraphics[width=0.5\textwidth]{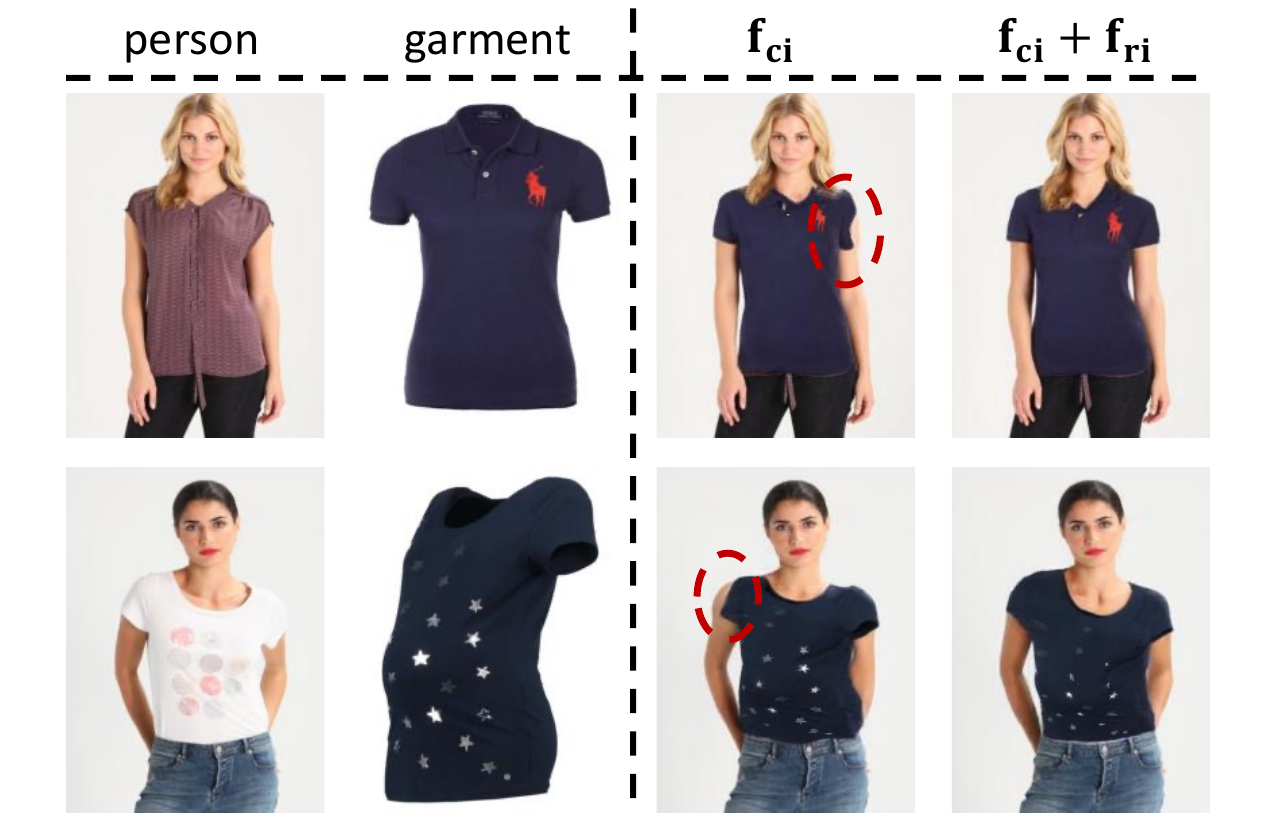}
    \vspace{-0.7cm}
    \caption{Comparing results with only $\mathbf{f_{ci}}$ used in $\mathcal{W}_{i}$ and $\mathbf{f_{ci}}+ \mathbf{f_{ri}}$ used in $\mathcal{W}_{i}$.}
    \vspace{-0.2cm}
    \label{fig:fig4}
\end{figure}

\begin{figure}[h!]
    \centering
    \includegraphics[width=0.5\textwidth]{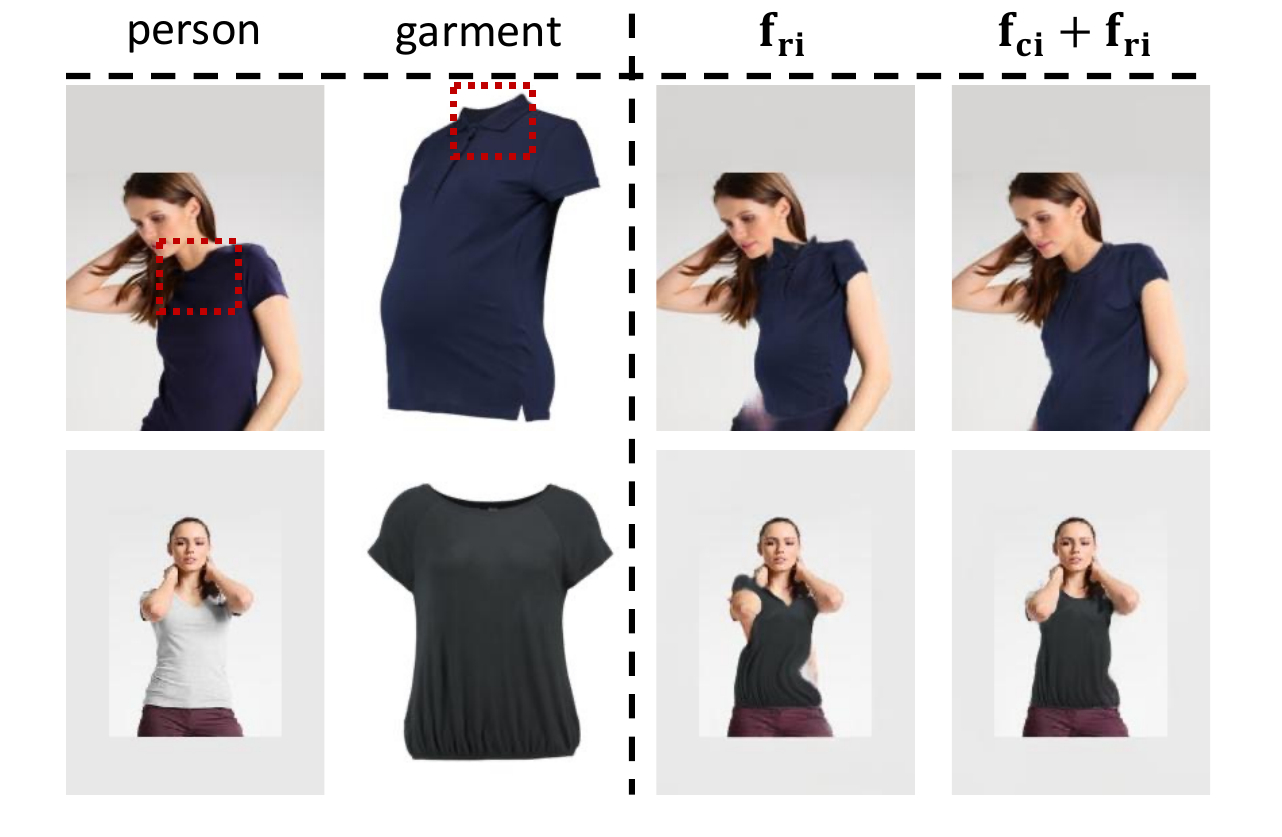}
    \vspace{-0.5cm}
    \caption{Comparing results with only $\mathbf{f_{ri}}$ used in $\mathcal{W}_{i}$ and $\mathbf{f_{ci}}+ \mathbf{f_{ri}}$ used in $\mathcal{W}_{i}$ in the case of large misalignment between the input person image and garment image.}
    \label{fig:fig5}
\end{figure}

\paragraph{Ablation Study} In this experiment, we validate the design of our appearance flow estimation blocks ($\mathcal{W}_{i}$). Specifically, we first experiment our method with only global style modulation (SM) based appearance flow estimation, that is, only using $\mathbf{f_{ci}}$ in Equation~\ref{eq:2} in each $\mathcal{W}_{i}$. We then experiment our method with only refinement flow (RF) estimation, that is, only using $\mathbf{f_{ri}}$ in Equation~\ref{eq:3} in each $\mathcal{W}_{i}$. Finally, we experiment with our combined method (SM + RF) which first estimates the appearance flow globally via style modulation and then refines the appearance flow locally through local correspondence. The quantitative results are shown in Table~\ref{tab:tab4}. Our proposed global style modulation (SM) based appearance flow method outperforms the local correspondence based method. When they were combined, the performance is further boosted. As illustrated in Fig.~\ref{fig:fig4}, without local refinement, our method (global style modulation only) sometimes cannot accurately predict the local fine-grained appearance flow, e.g., the sleeve regions, and thus generates unsatisfactory try-on image. However, with only local correspondence based appearance flow estimation, e.g., only using $\mathbf{f_{ri}}$ in $\mathcal{W}_{i}$, the method suffers when the corresponding regions are not located in the same receptive field. As illustrated in Fig.~\ref{fig:fig5}, $\mathbf{f_{ri}}$ cannot accurately estimate the appearance flow when there exists a large misalignment between the input person images and garment images. Once $\mathbf{f_{ci}}$ was first used to reduce the misalignment, our model can successfully overcome the problem. 

\begin{table}[ht]
    \centering
    \begin{tabular}{c|c|c}
    \toprule
         Methods & SSIM $\uparrow$& FID$\downarrow$\\
         \midrule
         RF& 0.89& 10.73\\
         SM& 0.89& 9.84\\
         \midrule
         SM + RF & \textbf{0.91}& \textbf{8.89}\\
         \bottomrule
    \end{tabular}

    \caption{Results on VTON testing dataset  when different appearance flow estimation methods were used in $\mathcal{W}_{i}$. RF: local correspondence based flow estimation. SM: style modulation based flow estimation.}
    \label{tab:tab4}
    \vspace{-0.4cm}
\end{table}

\section{Conclusion}
In this paper, we have proposed a style based global appearance flow estimation method to warp the garment for virtual try-on. Our method via style modulation first estimates the appearance flow globally and then refines the appearance flow locally. Our method achieves state-of-the-art performance on the VITON benchmark and it is more robust against large mis-alignment between person and garment images, as well as difficult poses/occlusions. We conducted extensive experiments to show the superiority of our method and validated our architecture design.

%%%%%%%%% REFERENCES
{\small
\bibliographystyle{ieee_fullname}
\bibliography{egbib}
}

\end{document}